\documentclass[10pt,twocolumn,letterpaper]{article}

\usepackage{cvpr}
\usepackage{times}
\usepackage{epsfig}
\usepackage{graphicx}
\usepackage{amsmath}
\usepackage{amssymb}
\usepackage{bm}
\usepackage{multirow}
\usepackage{array}

\usepackage[breaklinks=true,bookmarks=false]{hyperref}

\cvprfinalcopy 

\def\cvprPaperID{****} 

\begin{document}

\title{On Study of the Reliable Fully Convolutional Networks with Tree Arranged Outputs (TAO-FCN) for Handwritten String Recognition}

\author{Song Wang, Jun Sun, Satoshi Naoi, \\
{Fujitsu Research \& Development Center, Beijing, China}\\
song.wang.cn@outlook.com\\
\{wang.song, sunjun, naoi\}@cn.fujitsu.com\\
}

\maketitle

\begin{abstract}
The handwritten string recognition is still a challengeable task, though the powerful deep learning tools were introduced. In this paper, based on TAO-FCN, we proposed an end-to-end system for handwritten string recognition. Compared with the conventional methods, there is no preprocess nor manually designed rules employed. With enough labelled data, it is easy to apply the proposed method to different applications. Although the performance of the proposed method may not be comparable with the state-of-the-art approaches, it's usability and robustness are more meaningful for practical applications.
\end{abstract}

\section{Introduction}
\label{intro}
The handwritten string recognition is a very difficult task even for the cutting edge deep learning approaches. Its difficulty may reveal from the following aspects. First, the appearance of the character varies widely according to different writers. Besides, some specific language may have large class number, such as Chinese and Japanese. Therefore, a powerful classifier is necessary for handwritten recognition. Second, in practical application, there is various background noise and degradation of the characters. This requires high robustness of the method. Third, before the recognition process, the character string should be first detected and then extracted from the input image. This also increases the difficulty of the handwritten recognition to a large extent. In conclusion, we still have to face many challenges of the handwritten string recognition before we put it into practice.\par

There are many trials which focus on the handwritten string recognition. Among all these trials, the Long Short-Term Memory recurrent neural network (LSTM) is seen as the optimal choice. For example, in~\cite{4531750} the LSTM was applied to handwritten Latin string recognition and achieved very good performance. However in~\cite{messina_segmentation-free_2015}, when LSTM was applied to handwritten Chinese recognition, it only got about 90\% accuracy rate. This performance is not so different with the conventional methods in~\cite{qiu-feng_wang_handwritten_2012}. This is because it is very hard to train an LSTM model with a large number of output classes. \par

In order to solve the large class number problem, in~\cite{7814044} the powerful convolutional neural networks (CNN) was introduced to handwritten Chinese recognition. It is well known that among the deep learning models, CNN is the optimal choice for image classification tasks. This is because CNN has powerful classification ability. For example, in~\cite{7486592} the CNN achieved an even better performance than human on the task of isolated handwritten Chinese character recognition (3,755 classes in total). Nevertheless, since CNN is not able to generate reliable probabilities for the recognition results, it is not suitable for handwritten string recognition. In~\cite{7814044} this problem was solved by the heterogeneous CNN structure and which was able to generate reliable probabilities for the final path search process of the string recognition. As a result, the heterogeneous CNN had about 5\% improvement compared with the LSTM and the conventional best method. Although the heterogeneous CNN had very impressive performance, it is still not the optimal choice for handwritten string recognition. The reason is that the heterogeneous CNN needs manually designed rules for over-segmentation of the character string and which is sensitive to noise and degradation of the character image. In contrast, LSTM is more robust since its input is just the original image data.\par

Besides the LSTM and CNN, there is also another deep learning approach which can be considered for handwritten string recognition: the fully convolutional networks (FCN). FCN is originally designed for semantic segmentation of images~\cite{Long_2015_CVPR,DBLP:journals/corr/ChenPKMY14}. In~\cite{Zhang_2016_CVPR} it was used for text line extraction from scene images. The advantage of FCN is that it can do the detection, extraction and recognition at the same time. In other words, the FCN can be seen as an end-to-end solution for object recognition. Inspired by this, if we treat the handwritten character as special ``object'', then the FCN can be seen as an end-to-end solution for handwritten string recognition.\par

\begin{figure}
\centerline{\includegraphics[width=0.45\textwidth]{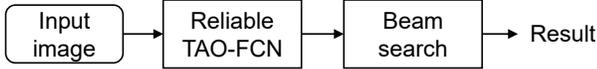}}
\caption{The process of the proposed method.} \label{process}
\end{figure}

In this paper, we proposed the reliable fully convolutional networks with tree arranged outpus (TAO-FCN), which is a combination of FCN and heterogeneous CNN. As shown in Fig.~\ref{process}, first, the TAO-FCN is used to recognize each pixel (with a certain area around) of the image as some character. Moreover, since it has the similar structure with the heterogeneous CNN, it can generate reliable probabilities. Second, a path search process of beam search is conducted based on the recognition results of the TAO-FCN. Finally, the recognition of the handwritten string is obtained. The merits of the proposed method are shown as follows.
\begin{itemize}
\item Since FCN is used instead of the manually designed rules and preprocess, the proposed method will be more robust and which can be easily applied to different tasks with enough training samples.\par
\item Compared with the conventional FCN, the TAO-FCN has better classification performance. As mentioned above, the classification ability of the classifier is important for handwritten string recognition.\par
\item Since the proposed method also has the path search process, it can inherit the advantages of the heterogeneous CNN, such as the deep integration with the language model\par
\end{itemize}
\par

The rest of the paper is organized as follows. The reliable TAO-FCN is introduced in Section~\ref{TAO-FCN} and the whole process of the proposed method is introduced in Section~\ref{Methodology}. In Section~\ref{experiments}, both the classification ability of the TAO-FCN and the performance of the whole method are analyzed. Finally, a conclusion is given in Section~\ref{Conclusion}. \par

\section{The Reliable Fully Convolutional Networks with Tree Arranged Outputs}
\label{TAO-FCN}

The original FCN is designed for object detection and extraction. It's purpose is to find the exact edges of different objects. Therefore, in FCN, the size of the original image is first decreased by the pooling operation and then recovered by the decovolutional operation. Through the above process, the edges of the objects may maintain clearly. However, the information inside the objects may be lost since the pooling operation discards the units of the feature maps. This information loss may bring negative influence to the classification power of the model.\par

\begin{figure}
\centerline{\includegraphics[width=0.4\textwidth]{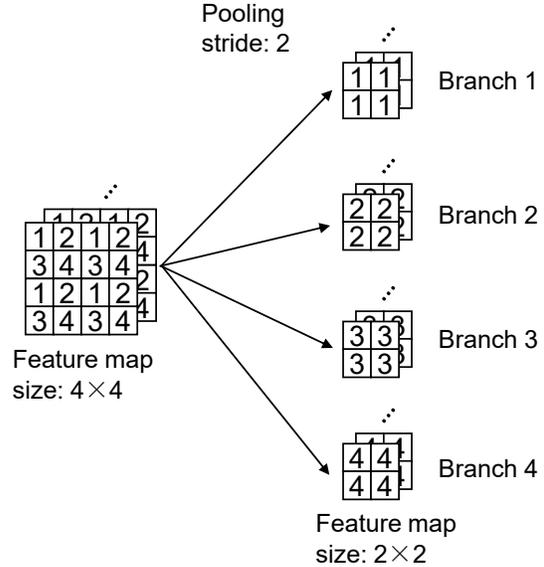}}
\caption{The pooling of the TAO-FCN.} \label{pooling}
\end{figure}

In order to avoid the information loss caused by the pooling operation, we proposed the TAO-FCN. As shown in Fig.~\ref{pooling}, the pooling layer of TAO-FCN will generate the branches to maintain all the units of the feature maps. For example, in Fig.~\ref{pooling}, if the stride of the pooling layer is 2, then each feature map will be divided into four sub-maps and each sub-map belongs to a network branch. By the way, the size of the sub-map is half of the original feature map. By using this special pooling layer, the TAO-FCN will not discard any unit of the network.\par

\begin{figure*}
\centerline{\includegraphics[width=0.8\textwidth]{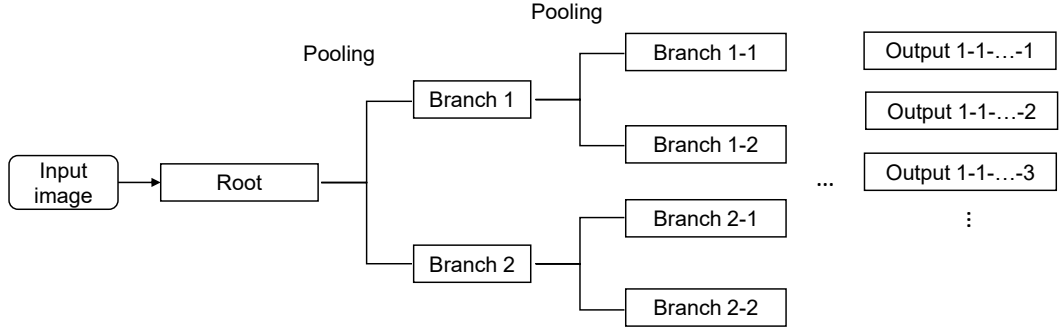}}
\caption{The structure of the TAO-FCN.} \label{wholenet}
\end{figure*}

The structure of the whole TAO-FCN network is shown in Fig.~\ref{wholenet}. After each pooling operation, the network will be divided into branches and even branch's branch. The more pooling layers used, the more branches generated. This is why the TAO-FCN is called the ``tree arranged outputs''. Obvious, there will be many output branches of the TAO-FCN and for each output branch, we must prepare a label. The deconvolutional layer is not used in TAO-FCN since there is no information loss. If we want an output result which has the same size with the input image, we just need to put all the results of the branches together to create a single output result. This combination result will have the same size with the input.\par

In TAO-FCN, if we just focus on one output branch, then it can be seen as a sub-network. Clearly, this sub-network is just a normal network with the normal pooling operation. Although this sub-network ``discards'' many units of the feature maps, these units are maintain by other branches. \par

\begin{figure*}
\centerline{\includegraphics[width=0.95\textwidth]{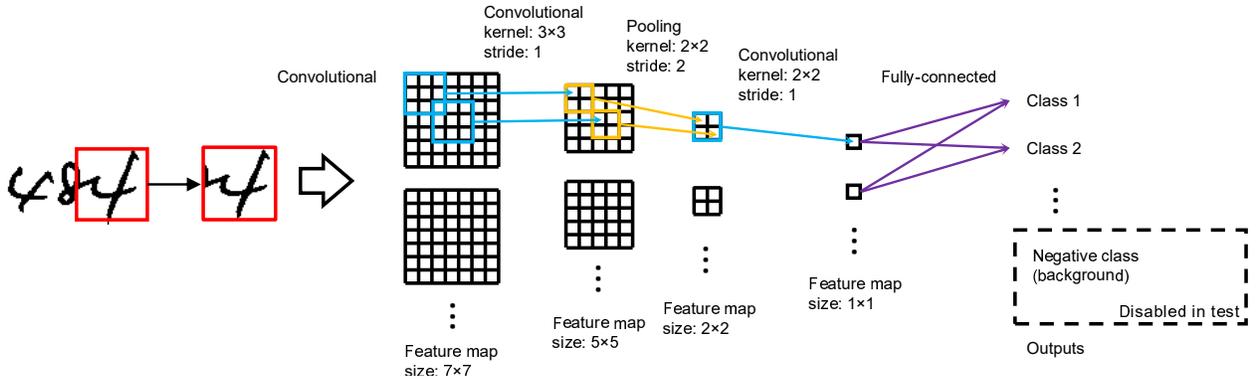}}
\caption{The demonstration of the single patch recognition process.} \label{asCNN}
\end{figure*}

Actually, to our understanding, the TAO-FCN just recognizes the patches extracted from the input image. The patches are extracted pixel by pixel. In other words, each pixel of the input image is seen as the center of a patch of a certain size and then this patch is recognized by TAO-FCN. As shown in Fig.~\ref{asCNN}, for a certain patch of the input image of TAO-FCN, it is processed just like the normal CNN with the convolutional, pooling and fully-connected layers. The difference is that, there is no padding operation in the convolutional layer. This is because the convolutional calculation results of the overlapping area between different patches are shared by these patches and the padding operation will break this sharing. Please note that for the whole TAO-FCN the padding can be used. This is because such padding is added at the boundaries of the input image and there is no sharing at the boundaries. \par

The size of the patch of the image is determined by the convolutional operation number and pooling operation number of TAO-FCN. For example, as shown in Fig.~\ref{asCNN}, a convolutional operation of $3\times3$ kernel will decrease the patch size by 2 and after each pooling layer (if pooling stride is 2) the size will be reduced by half. Besides, in TAO-FCN, the fully-connected layer for the patch is realized by the convolutional operation with $1\times1$ kernel.\par

In conclusion, the process of TAO-FCN is equal to use a normal CNN to recognize each patch of the input image. Because of the sharing of the convolutional calculation results, the TAO-FCN uses much less time than the normal CNN on recognizing all the patches. Besides, as shown in Fig.~\ref{asCNN}, in order to obtain reliable probabilities, the output of each patch is designed as the same as the heterogeneous CNN in~\cite{7814044}.\par

\section{Methodology}
\label{Methodology}

\begin{figure*}
\centerline{\includegraphics[width=0.95\textwidth]{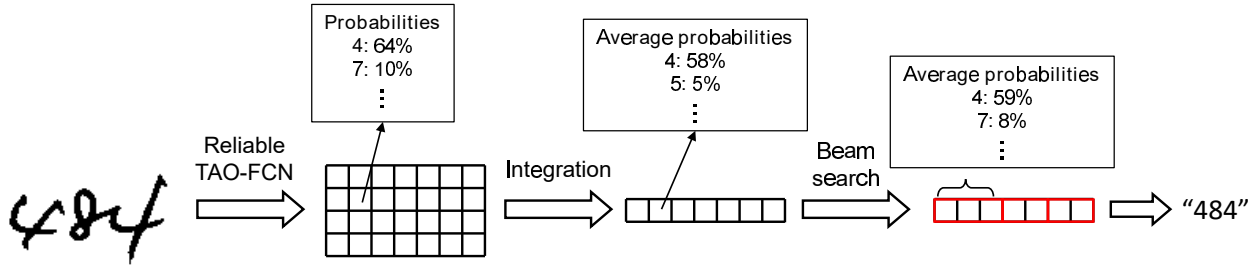}}
\caption{The framework of the proposed method. Each box stands for the recognition results of a certain pixel (patch).} \label{flowchart}
\end{figure*}

As shown in Fig.~\ref{flowchart}, the proposed method mainly has three modules: the recognition of reliable TAO-FCN, the integration of the probabilities and beam search for the final result. First, the input image is recognized by the reliable TAO-FCN and then the reliable probabilities are generated for each pixel. second, all the probabilities are integrated as one dimensional data and which will be the input of the beam search. Finally, by using the beam search, the final result is obtained.\par

\subsection{recognition by reliable TAO-FCN}
In this step, the input image is recognized by the reliable TAO-FCN. Each pixel of the input image is recognized as some character with the corresponding probabilities. As mentioned above, since the reliable TAO-FCN has the similar structure with the heterogeneous CNN, its generated probabilities are reliable. This is also the basis of the beam search (which requires accurate probabilities to score each path). \par

Obviously, because every pixel of the input image is recognized, the preprocess and noise removal are not necessary anymore. Simply by using the recognized probabilities of each pixel, we can distinguish the noise pixel and the character pixel. Furthermore, the degradation of the character can also be solved by learning from the training samples. That is to say, with enough training samples of various noise and degradation, we can train a robust reliable TAO-FCN.\par

\subsection{integration of the probabilities}
As shown in Fig.~\ref{flowchart}, after the probabilities of the pixels are obtained, the next step is to integrate the pixels into an one dimensional series. This is because the beam search can only accept one dimensional input. The integration is realized simply by the average probabilities of the pixels in a vertical line. These pixels will become one pixel in the final series. \par

\subsection{beam search}
After the one dimensional series is obtained, the next step is to apply beam search to find the optimal path. As shown in Fig.~\ref{flowchart}, in beam search, different combinations of the pixels are evaluated and then the combinations with the highest probabilities are seen as the final result. The probabilities for a certain combination of the pixels is simply calculated as the average probabilities of these pixels.\par

\section{Experimental Analysis}
\label{experiments}

In this section, we have two parts of the experiments: the evaluation of the isolated character recognition and the evaluation of the character string recognition. The evaluation of the isolated character recognition can show the classification ability of the TAO-FCN (compared with the normal FCN and heterogenous CNN). The string recognition is the evaluation of the whole proposed method.\par

\begin{figure}
\centerline{\includegraphics[width=0.3\textwidth]{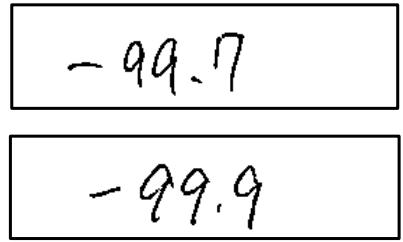}}
\caption{Examples of the dataset of the handwritten digit strings.} \label{examples}
\end{figure}

As shown in Fig.~\ref{examples}, in the experiments, an internal dataset of the handwritten digit string was used with two extra symbol classes (minus ``$-$'' and dot ``$.$''). This dataset includes the specific segmentation label of each character, thus it can be used to train the reliable TAO-FCN. The selection of the dataset is based on the following considerations. First, although the handwritten Chinese string may be a good choice, a TAO-FCN for Chinese characters will cost much memory since the class number of which is very large. Currently our computational resource is not enough for the Chinese characters. Second, it is very difficult to find a handwritten Latin dataset with segmentation labels. Finally, we decided to use our internal dataset of digits with the segmentation labels.\par

In the dataset, there are 5,398 strings for training and 469 strings for test. As mentioned above, the total class number is 12, including 10 digits and 2 symbols. \par

\begin{figure*}
\centerline{\includegraphics[width=0.7\textwidth]{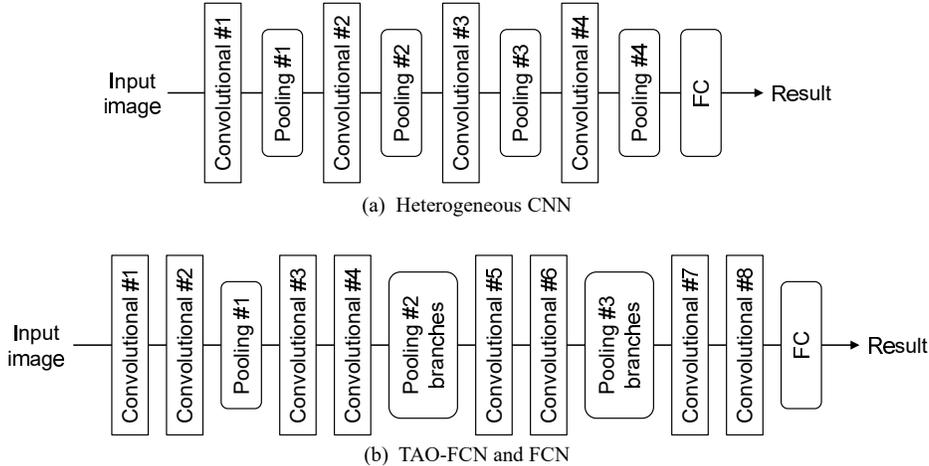}}
\caption{Network structure of the three models.} \label{models}
\end{figure*}

The structure of the three models are shown in Fig.~\ref{models}. For heterogeneous CNN, a four-layer network was used with four pooling operations. For TAO-FCN and FCN, an eight-layer network was used with three pooling operations. In order to save memory of the GPU, the branching starts from the second pooling operation. Obviously, the structure of the TAO-FCN and FCN is larger than heterogeneous CNN. This is because the recognition target of the TAO-FCN and FCN is much more complicated than the heterogeneous CNN. \par

\subsection{evaluation on isolated character recognition}

\begin{figure}
\centerline{\includegraphics[width=0.35\textwidth]{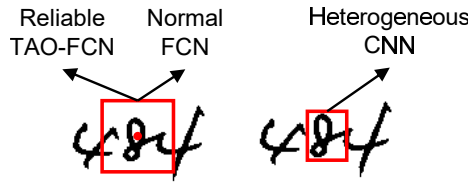}}
\caption{The isolated character recognition result of the three models.} \label{isolated}
\end{figure}

First, the training set was used to train the reliable TAO-FCN, normal FCN (without deconvolutional layer) and heterogeneous CNN. Afterwards, each test string was recognized by the three models. We assume that the bounding box of each character in the test string is known and then we calculate the recognition rate of the characters. In other words, this recognition rate is calculated based on the correct segmentation of each character. Since this isolated character recognition rate is from the real character string, it can reflect the classification ability of the models in handwritten string recognition. As shown in Fig.~\ref{isolated}, for reliable TAO-FCN and FCN, the recognition result of the center pixel of the character was seen as the final recognition result while for the heterogeneous CNN the final recognition result was from the character of the correct segmentation. \par

The results of the three models are shown in Table~\ref{isolatedExp}. First, the heterogeneous CNN had the highest recognition rate. This is because the over-segmentation process removed most of the unreasonable patches and as a result, the difficulty of the classification was reduced. However, in TAO-FCN and FCN, since each patch of the input image was recognized, the classification difficulty is relatively high. Second, the TAO-FCN achieved better performance than the normal FCN. This proved that without the information loss, the TAO-FCN is able to have better classification ability. Third, we have to say the performance of TAO-FCN is still a little bit lower than the CNN on the classification. However, if we continue to enlarge the network structure, the classification ability of the TAO-FCN can be further improved.\par

\begin{table}[ht]
\caption{Experiment results on isolated character recognition.}
\vspace{0.5\baselineskip}
\centering
{\begin{tabular}{c|c}
Method & recognition rate (\%) \\ \hline
FCN & 96.8   \\
TAO-FCN & 98.1  \\
Heterogeneous CNN & 98.6 \\
\end{tabular}
}
\label{isolatedExp}
\end{table}

\subsection{evaluation on the string recognition}

The accuracy rate (AR) and correct rate (CR) (defined in~\cite{7814044}) of the proposed method and heterogeneous CNN are shown in Table~\ref{stringexp}. Through the results we can see that on the string recognition the performance of TAO-FCN was comparable with the heterogeneous CNN. The CR of TAO-FCN was 1\% higher than the heterogeneous CNN. This proved that the TAO-FCN is able to achieve the same level performance with the state-of-the-art method on handwritten string recognition. \par

\begin{table}[ht]
\caption{Experiment results on string recognition.}
\vspace{0.5\baselineskip}
\centering
{\begin{tabular}{c|c c }
Method & CR (\%) & AR (\%)\\ \hline
Proposed method & 97.1 & 93.6  \\
Heterogeneous CNN & 96.1 & 93.9 \\
\end{tabular}
}
\label{stringexp}
\end{table}

\section{Conclusion and Future Work}
\label{Conclusion}

In this paper, the reliable TAO-FCN is proposed for handwritten string recognition. Compared with other solutions, the proposed method is supposed to be robust and easy to use. We can simply train the TAO-FCN with enough samples for different tasks without using manually designed rules. In the experiments, the TAO-FCN achieved almost the same performance of the heterogeneous CNN on isolated character recognition. This result shows the potential of the TAO-FCN. Moreover, the string recognition rate of the TAO-FCN was also comparable with the state-of-the-art. \par

For the future work, first we will continue to improve the classification ability of the TAO-FCN, which is the basis for the whole solution. Besides, we will try other score function of the beam search to improve the final performance of the proposed method.\par

{\small
\bibliographystyle{ieee}
\bibliography{References}
}

\end{document}